\documentclass[letterpaper, 10 pt, conference]{ieeeconf}  

\IEEEoverridecommandlockouts                              

\usepackage{graphics} 
\graphicspath{ {./images/} }
\usepackage{subcaption}

\usepackage{amsmath} 
\DeclareMathOperator*{\argmax}{argmax}
\usepackage{amssymb}  
\usepackage{algorithm}
\usepackage[noend]{algpseudocode}
\usepackage{lipsum}
\usepackage{gensymb}

\usepackage{amsthm}
\usepackage{float}
\newtheoremstyle{mystyle}
  {}
  {}
  {\itshape}
  {}
  {\bfseries}
  {.}
  { }
  {}
\theoremstyle{mystyle}
\newtheorem{definition}{Definition}

\newtheorem{assumption}{Assumption}
\usepackage{todonotes} 

\title{\LARGE
Collision-Aware Target-Driven Object Grasping in Constrained Environments
}
\author{Xibai Lou$^{1}$, Yang Yang$^{2}$ and Changhyun Choi$^{1}$
\thanks{*This work was in part supported by the MnDRIVE Initiative on Robotics, Sensors, and Advanced Manufacturing.}
\thanks{$^{1}$X. Lou and C. Choi are with the Department of Electrical and Computer Engineering, Univ. of Minnesota, Minneapolis, USA
        {\tt\small \{lou00015, cchoi\}@umn.edu}}%
\thanks{$^{2}$Y. Yang is with the Department of Computer Science and Engineering, Univ. of Minnesota, Minneapolis, USA {\tt\small yang5276@umn.edu}}%
}

\begin{document}

\maketitle
\thispagestyle{empty}
\pagestyle{empty}

\begin{abstract}
Grasping a novel target object in constrained environments (e.g., walls, bins, and shelves) requires intensive reasoning about grasp pose reachability to avoid collisions with the surrounding structures. Typical 6-DoF robotic grasping systems rely on the prior knowledge about the environment and intensive planning computation, which is ungeneralizable and inefficient. In contrast, we propose a novel Collision-Aware Reachability Predictor (CARP) for 6-DoF grasping systems. The CARP learns to estimate the collision-free probabilities for grasp poses and significantly improves grasping in challenging environments. The deep neural networks in our approach are trained fully by self-supervision in simulation. The experiments in both simulation and the real world show that our approach achieves more than 75\% grasping rate on novel objects in various surrounding structures. The ablation study demonstrates the effectiveness of the CARP, which improves the 6-DoF grasping rate by 95.7\%.
\end{abstract}
\smallbreak

\section{INTRODUCTION}

\begin{figure}[t]
    \includegraphics[scale=0.40]{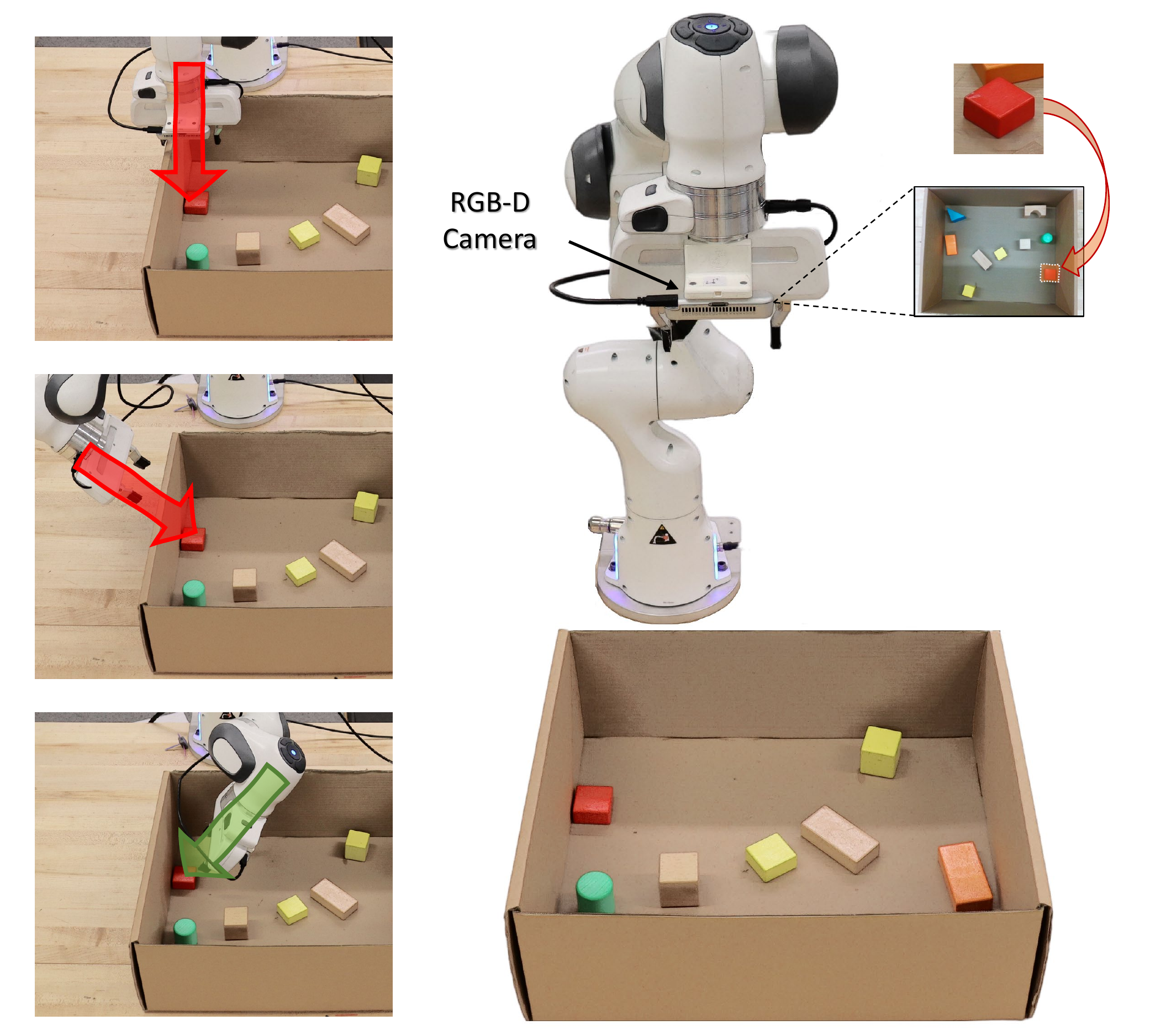}
    \caption{\textbf{Grasping a target object in constrained environments.} Given a query image, our approach is able to localize and grasp the target object (red cuboid) surrounded by structures through reasoning feasible 6-DoF poses. In particular, our Collision-Aware Reachability Predictor (CARP) helps solve this challenging task by estimating the collision-free probability of each grasp pose.}
  \label{fig:cover}
  \vspace{-8pt}
\end{figure}

Target-driven grasping is a fundamental yet challenging task in robotic manipulation, as it requires intensive reasoning about grasping stability from imperfect and partial observations. Most grasping systems assume a table-top scenario and simply choose 3-DoF grasp poses to mitigate the difficulty of reasoning. However, to grasp novel targets in constrained environments, an autonomous robot has to expand its action space from 3-DoF to 6-DoF, as shown in Fig. \ref{fig:cover}. In addition, these environments escalate two challenges: 1) How to robustly perceive novel target objects and surrounding structures? and 2) How to foresee the influence of surrounding structures on the grasping success probability?

In recent years, target-driven grasping approaches have been proposed by combining off-the-shelf object recognition modules (e.g., detection, template matching, and classifiers) with data-driven grasping models~\cite{fang2018multi}, \cite{8584228}. These approaches focus on object-centric reasoning for grasping (i.e., predicting grasping stability from object appearance or geometry) while overlooking scene context beyond objects. As a result, in constrained environments with surrounding structures (e.g., walls, bins), they have to plan for an enormous set of sampled grasp poses and iteratively search through the entire set with a collision-checking algorithm. Furthermore, these approaches require complete knowledge about the environment (e.g., geometric models of surrounding structure), which is, in practice, usually partially observable by imperfect sensors. Hence, these approaches suffer from excessive planning failures due to the absence of collision-awareness. Another limitation of these approaches lies in perception. Though simulation can expedite development and training, these RGB image-based object recognition modules require additional efforts to bridge the sim-to-real gap~\cite{closingsim2real}, \cite{bousmalis2018using} and generalize poorly to novel objects. These limitations motivate us to develop a target-driven grasping pipeline that achieves single-shot recognition for novel objects and requires only single planning for 6-DoF grasping in constrained environments. 

The proposed collision-aware target-driven grasping pipeline integrates a robust perception module and a collision-aware 6-DoF grasping module. The perception module exploits the depth information in simulation with Siamese networks~\cite{Koch2015SiameseNN} for single-shot recognition and sim-to-real generalization. Our 6-DoF grasping module features a Collision-Aware Reachability Predictor (CARP). The CARP is a 3D convolutional neural network (3D CNN) that explicitly learns the probabilities of reaching a set of grasp poses without having collisions between the robot manipulator and surrounding structures. The overall feasibility of 6-DoF grasp poses is evaluated by combining the collision-free probabilities and the predictions of grasping stability from a 3D CNN-based Grasp Stability Predictor (GSP) \cite{lou2020learning}.\footnote{Note that the acronym GSP refers to the 3D CNN module in~\cite{lou2020learning} We did not explicitly use the GSP in~\cite{lou2020learning} but defined here for a concise reference.}.

The rationale behind our approach resides in human behavior. For instance, when grasping an item from a packaging box, we naturally optimize our grasping action (in terms of both reachability and stability) with our experience rather than iterative checking. Our work is an early attempt to explicitly estimate collisions in constrained environments for target-driven 6-DoF grasping. The main contributions of our work are as follows:

\begin{itemize}
\item 
The Collision-Aware Reachability Predictor (CARP) that models the correlation between the spatial information and the collision-free probability of 6-DoF grasp poses with a 3D CNN. The CARP is trained with synthetic depth data by self-supervision and directly transferred to the real world. 
\end{itemize}

\begin{itemize}
\item 
A target-driven robotic grasping pipeline that comprises a depth-based single-shot recognition module, the Collision-Aware Reachability Predictor, and the Grasp Stability Predictor. The pipeline localizes the novel target object in clutter and then grasps it in a constrained environment with single planning.
\end{itemize}

\begin{figure*}[t]
\centering
\includegraphics[width=1.0\textwidth]{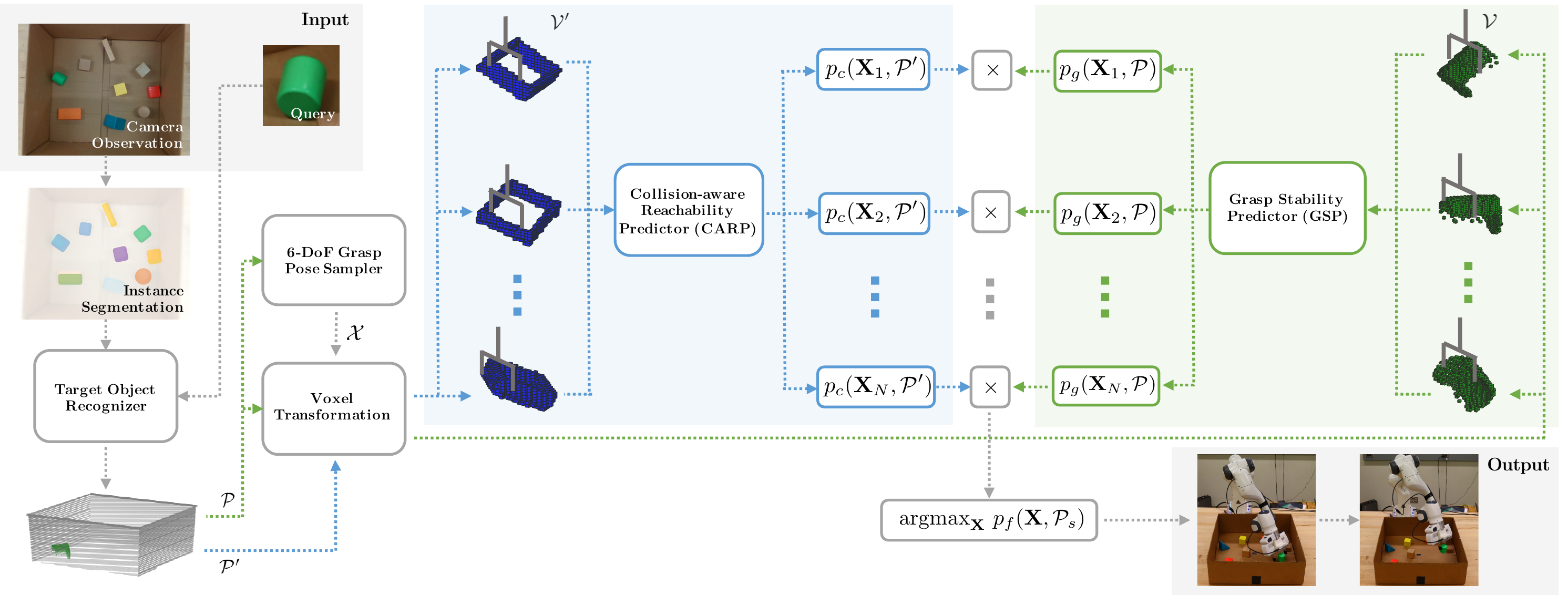}
\caption{\textbf{Grasping pipeline.} The scene point cloud $\mathcal{P}_s$ is first reconstructed from RGB-D image $\mathbf{I}$. Then the perception module takes $\mathbf{I}$ as input and gives the target object point cloud $\mathcal{P}$ and the surrounding structure cloud $\mathcal{P'}$. From $\mathcal{P}$, a set of 6-DoF grasp poses $\mathcal{X}$ is randomly sampled, and $\mathcal{P'}$ is transformed by each $\mathbf{X} \in \mathcal{X}$ and voxelized to voxel grid $\mathcal{V'}$. The CARP takes $\mathcal{V'}$ as input and evaluates the collision-free probability $p_c$. The voxel grid $\mathcal{V}$ is transformed from the object point cloud $\mathcal{P}$ w.r.t $\mathbf{X}$. The Grasp Stability Predictor then evaluates the grasping stability $p_g$ for each $\mathcal{V}$. We multiply $p_c$ and $p_g$ to get a grasping feasibility $p_f$ and execute the pose having the highest $p_f$.}
\label{fig:flowchart}
\vspace{-8pt}
\end{figure*}

\section{RELATED WORK}
\subsection{Target-driven Object Grasping}
Robotic grasping is a fundamental but challenging problem in robotic manipulation~\cite{sahbani2012overview}, \cite{bohg2013data}. This vast literature can be divided into model-based~\cite{miller2004graspit, weisz2012pose} and learning-based approaches~\cite{mahler2017dex}, \cite{levine2018learning}. Another way of categorization is by task goal: target-agnostic~\cite{pinto2016supersizing},\cite{zeng2018learning} and target-driven grasping~\cite{8584228},\cite{danielczuk2019mechanical}. Our approach is learning-based and target-driven. Recent target-agnostic grasping approaches apply deep neural networks to learn grasping in 3-DoF action space (i.e., grasp pose with a 2D position and a wrist orientation)~\cite{pinto2016supersizing}, \cite{zeng2018learning}, \cite{liang2019knowledge}, \cite{lenz2015deep}, \cite{Kappler2015LeveragingBD}. While 3D CNN was mainly studied in object recognition tasks \cite{maturana2015voxnet, song2017semantic}, Choi \emph{et al.}~\cite{choi2018learning} showed with a soft robot hand that it is capable of proposing additional grasping directions beyond the standard top-down poses. A few works learn to propose 6-DoF poses \cite{lou2020learning, gualtieri2016high, ten2017grasp, mousavian2019graspnet, song2020grasping}. Mousavian \emph{et al.} trained a network based on PointNet++ and a variational autoencoder \cite{Kingma2014} to generate stable grasp poses. Target-driven grasping \cite{8584228}, \cite{yang2020deep}, \cite{fang2020learning}, \cite{murali2020clutteredgrasping} is less studied compared to target-agnostic ones. The target-driven grasping problem necessitates a perception module, such as 2D image based template matching~\cite{8461041} and semantic  segmentation~\cite{yang2020deep}, that recognizes the target object before grasping. Single-shot recognition using traditional RGB-based Siamese Networks~\cite{Koch2015SiameseNN} has also been explored for robotic grasping task~\cite{danielczuk2019mechanical}. Recent studies suggests that synthetic depth data is less influenced by the sim-to-real gap~\cite{mahler2017dex}, \cite{danielczuk2019segmenting}. We exploit the depth data both for object perception and grasping to minimize such sim-to-real gap.

\subsection{Grasping Reachability}

Typical robotic grasping pipelines need to solve inverse kinematics with motion planning algorithms~\cite{LaValle00rapidly-exploringrandom}, \cite{kuffner2000rrt} to reach a goal pose. Though such algorithms can handle reachability and collision during execution, the computational cost is remarkably high, especially when the trajectory to execute is infeasible. Most approaches bypass this problem by restricting target objects within a known reachable workspace~\cite{zeng2018learning}, \cite{choi2018learning}, \cite{ten2017grasp}; other works estimate the grasping reachability by querying an offline \cite{reachability}, or online \cite{akinola2018workspace} database of reachable grasp poses. The Reachability Predictor in our prior work~\cite{lou2020learning} only predicts the reachability concerning the kinematics of robot arms in the table-top scenario. Hence, it does not consider the collision-free probability, which determines the grasping reachability in the constrained environments. A comparable work from Murali \emph{et al.}~\cite{murali2020clutteredgrasping} extends their previous work~\cite{mousavian2019graspnet} to estimate the collision score between object and gripper for 6-DoF grasp poses. We consider beyond gripper-object collisions as our collision sources further include surrounding structures.

\section{PROBLEM FORMULATION}
We consider the problem of generating feasible 6-DoF grasp poses for a target object surrounded by structures and other objects. The problem is formulated as follows:

\begin{definition}
\textit{A grasp pose $\mathbf{X}\in SE(3)$ is \textbf{collision-free} if the robot arm is able to reach the goal configuration without colliding with the surrounding structures.}
\end{definition}


\begin{assumption}
The target object is possibly unknown (i.e., novel objects) and partially observable (e.g., object occlusions and imperfect sensors), but an image of the target object is given as the only target information.
\end{assumption}


The scene point cloud $\mathcal{P}_s$, obtained from a single view RGB-D image $\mathbf{I}$, includes target object point cloud $\mathcal{P} \subset \mathbb{R}^3$ and a surrounding structure point cloud $\mathcal{P}' \subset \mathbb{R}^3$. To explore the full 6-DoF action space, we do not pose any constraint on the sampled grasp pose set $\mathcal{X}$. The detailed sampling procedure is described in \cite{lou2020learning}.

Let $\mathcal{S}_c(\mathbf{X}) \in \{0,1\}$ denote a binary-valued collision-free metric where $\mathcal{S}_c = 1$ indicates that the grasp is collision-free. The collision-free probability is determined solely by the spatial relationship between the manipulator and the surrounding structures, given by $p_c(\mathbf{X}, \mathcal{P}') = Pr(\mathcal{S}_c = 1|\mathbf{X}, \mathcal{P}')$.

Each grasp pose $\mathbf{X} \in \mathcal{X}$ is also subject to a binary-valued stability metric $\mathcal{S}_g(\mathbf{X})\in$ \{0,1\} where $\mathcal{S}_g = 1$ indicates that the grasp pose is stable. We would like to estimate the grasping stability $p_g(\mathbf{X}, \mathcal{P}) = Pr(\mathcal{S}_g = 1|\mathbf{X},\mathcal{P})$. Finally, the feasibility metric $\mathcal{S}_f(\mathbf{X}) \in$ \{0,1\} measures if the grasp pose is feasible to accomplish the task, and $\mathcal{S}_f = 1$ indicates a feasible, and therefore simultaneously stable and collision-free grasp pose. Note that the stability metric of a pose $\mathbf{X}$ is independent of its collision-free metric. Thus, we consider the grasping feasibility $p_f(\mathbf{X}, \mathcal{P}_s)$ as the joint probability of the two independent probabilities $p_c(\mathbf{X}, \mathcal{P}')$ and $p_g(\mathbf{X}, \mathcal{P})$.

\begin{algorithm}[b]
\caption{Perception Module}\label{algo:perception}
\hspace*{\algorithmicindent} \textbf{Input: }Scene RGB-D Image $\mathbf{I}$, Query RGB-D Image $\mathbf{I}_q$, SD Mask-RCNN $\mathcal{N}_s$, Siamese Network $\mathcal{N}_r$\\
\hspace*{\algorithmicindent} \textbf{Output: }Target Cloud $\mathcal{P}$, Structure Cloud $\mathcal{P'}$
\begin{algorithmic}[1]
\State $\mathcal{M} \gets \texttt{$\mathcal{N}_s$.InstanceSegmentation($\mathbf{I}$)}$
\For{${\mathbf{M}}\in\mathcal{M}$}
    \State $\mathbf{I}_o \gets \texttt{Mask}(\mathbf{I}, \mathbf{M})$
    \State $s(\mathbf{I}_o, \mathbf{M}) \gets \texttt{$\mathcal{N}_r$.Recognition($\mathbf{I}_o, \mathbf{I}_q$)}$
\EndFor
\State $\mathbf{I}_t, \mathbf{M}_t \gets \argmax_{\mathbf{I}_o} s(\mathbf{I}_o, \mathbf{M})$
\State $\mathbf{I}_s \gets {\mathbf{I}} \cap \overline{\mathbf{M}}_t$
\State $\mathcal{P, P'} \gets \texttt{BackProjection}(\mathbf{I}_t, \mathbf{I}_s)$
\end{algorithmic}
\end{algorithm}

\section{PROPOSED APPROACH}

We propose a 6-DoF target-driven pipeline for object grasping in constrained environments. Our approach aims to 1) recognize a target object (seen or novel) with a partial observation and 2) select the most feasible grasp pose from a set of randomly sampled 6-DoF grasp poses.

\subsection{System Overview}
As illustrated in Fig.~\ref{fig:flowchart}, our grasping pipeline comprises a perception module and a grasping module. Given a query image for the target object, the perception module first separates the scene point cloud $\mathcal{P}_s$ into target point cloud $\mathcal{P}$ and surrounding structures $\mathcal{P}'$. The point clouds are then forwarded to our grasping module, which consists of the Collision-Aware Reachability Predictor (CARP) and the Grasp Stability Predictor (GSP). The CARP evaluates the collision-free probability $p_c(\mathbf{X}, \mathcal{P}')$ for grasp pose candidates $\mathbf{X}$ by using the structure point cloud $\mathcal{P}'$. The CARP takes advantage of the spatial relationship between the robot hand and all the surrounding structures to determine if the given pose is prone to collision with the structure. The GSP evaluates the stability of the grasp poses on the object point cloud $\mathcal{P}$. The final grasp pose is selected to maximize the overall grasping feasibility $p_f(\mathbf{X}, \mathcal{P}_s)$. 
Our grasping pipeline can be seen as a combination of a peripheral vision (i.e., the CARP's coarse but wide-angle understanding of the surrounding structures) and a foveated vision (i.e., the GSP's finer yet narrow-viewed understanding of the target object).

\subsection{Perception}
The perception module is delineated in Algorithm~\ref{algo:perception} wherein it first takes as input a scene RGB-D image $\mathbf{I}$ from the camera and generates a set of binary-valued class-agnostic instance masks $\mathcal{M}_{1, \cdots, N} \in \mathbb{Z}^{H \times W}$ by SD Mask R-CNN~\cite{danielczuk2019segmenting}. Next, the set of the masks is multiplied with the RGB-D image to generate object images for recognition, where a queried RGB-D image of the target object is supplied. Unlike a traditional RGB image-based Siamese network~\cite{Koch2015SiameseNN}, our approach includes the depth channel that helps differentiate the challenging objects (e.g., having similar visual appearance but different geometric shapes) and improves the sim-to-real generalization. To achieve single-shot recognition, our RGB-D Siamese CNN first extracts the latent features of an object image and the queried image for feature matching. During training, the L1-distance between the two feature vectors is minimized for the same class objects using a contrastive loss~\cite{hadsell2006dimensionality}. Therefore, the best-matched object during testing will have the lowest distance in a forward pass and is selected as the target object. The masked depth image is used to reconstruct the 3D target point cloud $\mathcal{P}$ through back-projection while the non-masked region gives the structure point cloud $\mathcal{P}'$.
\begin{figure}[t]
    \centering
    \includegraphics[width=0.45\textwidth]{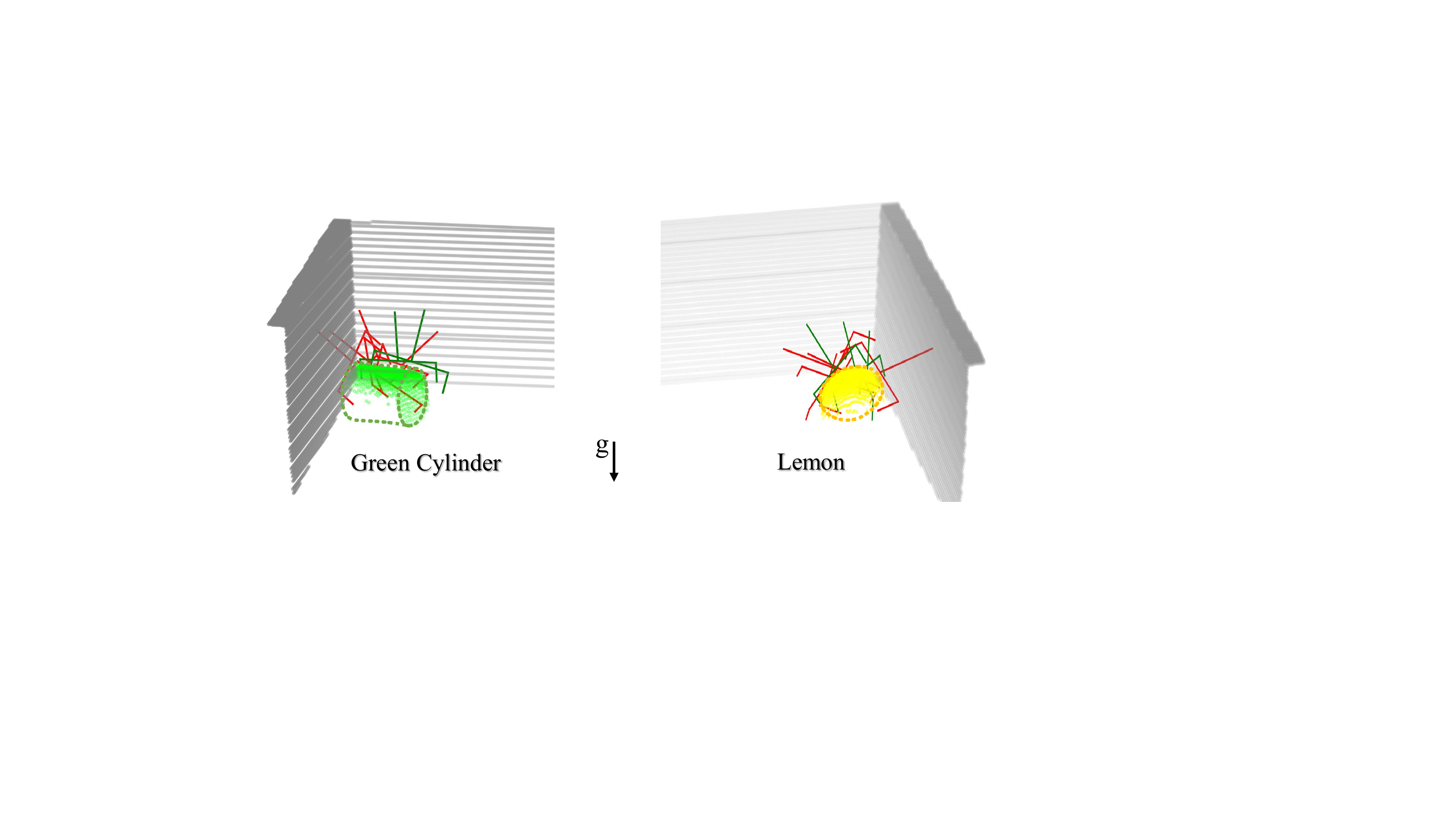}    
    \caption{\textbf{The CARP predictions.} The target object is placed adjacent to the surrounding structures. The CARP predicts the collision-free probabilities of each grasp pose. The green poses have higher results and are less prone to collision.}
    \label{fig:prediction}  
    \vspace{-8pt}
\end{figure}

\subsection{Grasping}

\begin{algorithm}[b]
\caption{Collision-Aware Target-driven Grasping}\label{algo:grasping}
\hspace*{\algorithmicindent} \textbf{Input: }RGB-D image $\mathbf{I}$, GSP  $\mathcal{N}_g$, CARP $\mathcal{N}_c$\\
\hspace*{\algorithmicindent} \textbf{Output: }collision-free grasp pose $\mathbf{X_f}\in{SE(3)}$ 
\begin{algorithmic}[1]
\State $\mathcal{P}_s \gets \texttt{BackProjection($\mathbf{I}$)}$
\State $\mathcal{P, P'} \gets \texttt{Perception($\mathbf{I}$)}$
\State $\mathcal{X} \gets \texttt{GraspPoseSampling($\mathcal{P}$)}$
\For{${\mathbf{X}}\in\mathcal{X}$}
    \State $\mathcal{P} \gets \texttt{CenterCropping($\mathcal{P}_s, \mathbf{X}$)}$
    \State $\mathcal{V}, \mathcal{V'} \gets \texttt{VoxelTransformation}(\mathcal{P}, \mathcal{P'}, \mathbf{X})$
    \State $p_c \gets \texttt{$\mathcal{N}_c$.Feedforward}(\mathcal{V'})$
    \State $p_g \gets \texttt{$\mathcal{N}_g$.Feedforward}(\mathcal{V})$
    \State $p_f \gets p_c \times p_g$
\EndFor
\State $\mathbf{X_f} \gets \argmax_{\mathbf{X} \in \mathcal{X}} p_f(\mathbf{X}, \mathcal{P}_s)$
\State $\texttt{Grasp}(\mathbf{X_f})$
\end{algorithmic}
\end{algorithm}

Our grasping module first considers the entire environmental structures to capture all potential collisions. It encodes the spatial information by transforming structure cloud $\mathcal{P}'$ with respect to the grasp pose $\mathbf{X}$. Then it voxelizes the transformed $\mathcal{P}'$ to a $40 \times 40 \times 40$ binary voxel occupancy grid $\mathcal{V}'$, whose voxel size is $(0.025m)^3$. It is centered on the grasp point $\mathbf{p}$, and its coordinate frame is aligned with that of the grasp pose $\mathbf{X}$. Consequently, collision-free and colliding poses correspond to $\mathcal{V}'$ with distinct characteristics, and the CARP learns from these features to estimate the collision-free probability $p_c$ from the input voxel. The spatial features are extracted by the 3D CNNs and then fed to fully connected layers. The output layer uses a sigmoid activation function to model the collision-free probability of $\mathbf{X}$. Since a given grasp pose can either be 0 (collision) or 1 (collision-free), we use binary cross-entropy as the loss function. 

To estimate the grasping stability for 6-DoF poses, we use a 3D CNN-based Grasp Stability Predictor (GSP), details of implementation can be found in~\cite{lou2020learning}. It focuses on the target object and examines the geometric shape. Moreover, it implicitly minimizes collisions with clutters by evaluating a more informative update of $\mathcal{P}$, obtained by center cropping the scene point cloud $\mathcal{P}_s$ at a grasping point $\mathbf{p}$ by $(0.1 m)^3$. The new $\mathcal{P}$ is transformed with respect to the grasp pose $\mathbf{X}$ and then voxelized as the input $\mathcal{V}$ to the GSP. Since $\mathcal{V}$ is voxelized from the cropped point cloud $\mathcal{P}$, it may partially contain the voxels of adjacent objects, which tend to lower grasping stability predictions as these voxels make the geometric shape in $\mathcal{V}$ unfamiliar. Therefore, the GSP will pick the pose that contains the least collision with other surrounding objects. We choose the most feasible pose as the final grasp pose to be executed, which is simultaneously collision-free and stable. Algorithm~\ref{algo:grasping} shows the flow of our grasping system.

\subsection{Data Collection and Training}
The entire system is trained by self-supervision in simulation. We first train the perception module with 500 RGB-D images of each object with ground truth class labels (accessible in simulation). For grasping module, we generated the training dataset by dropping the training objects within a common workspace structure (e.g., wall or bin) of random size (ranging from $0.3 m$ to $0.5 m$) and orientation. Then the robot interacts with the objects and collects 60,000 labeled point clouds to train the CARP. The labels are generated reliably and efficiently with the default collision checking algorithm in the simulation environment, which assumes the full knowledge of the scene. To decouple collision from grasping results, a training dataset of 50,000 data is collected separately to train the GSP in the table-top scenario where no surrounding structures exist.

\section{EXPERIMENTS}
We evaluate our approach in both simulated and real-world settings. The experiments are designed to answer three questions: 1) Can the perception module robustly identify the target object, including novel ones, given a query image?, 2) Can the CARP improve planning efficiency?, and 3) How does our approach perform compared to other state-of-the-art grasping approaches in various testing scenarios?

\textbf{Baselines:} We first performed ablation studies on the perception module. The performance of the pipeline is then compared with 4 baseline methods: 1) \textbf{\texttt{RAND}} randomly generates a grasp pose on the target object, 2) \textbf{\texttt{6GN}} is 6-DoF GraspNet \cite{mousavian2019graspnet} that learns to generate 6-DoF grasp poses with PointNet++\cite{qi2017pointnet} and a variational autoencoder, 3) \textbf{\texttt{VPG}} \cite{zeng2018learning} learns both pushing and grasping from RGB-D images, and we only evaluate the grasping part of this work by filtering the grasp Q-map with a target object mask, and 4) \textbf{\texttt{GSP+RP}} combines the Grasp Stability Predictor (GSP) with the Reachability Predictor (RP)~\cite{lou2020learning} that only considers the kinematic constraints of a robot arm when proposing feasible 6-DoF grasp poses. Note that the mask output from our perception module is input to each baseline as they were originally designed for target-agnostic grasping.

\begin{figure}[b]
  \begin{subfigure}{0.24\textwidth}
    \includegraphics[width=\textwidth]{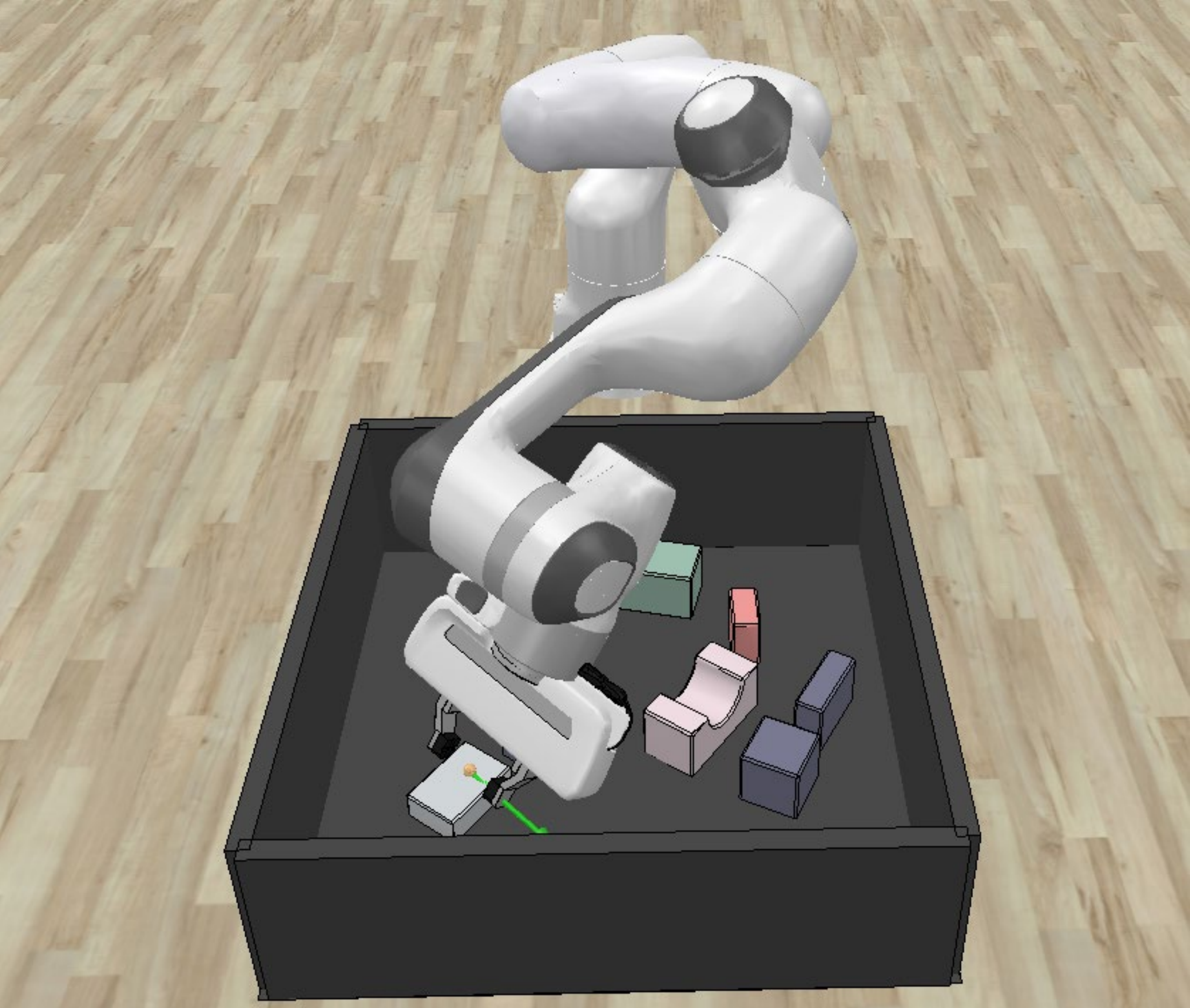}
    \caption{Standard arrangement}
  \end{subfigure}
  \hfill
  \begin{subfigure}{0.24\textwidth}
    \includegraphics[width=\textwidth]{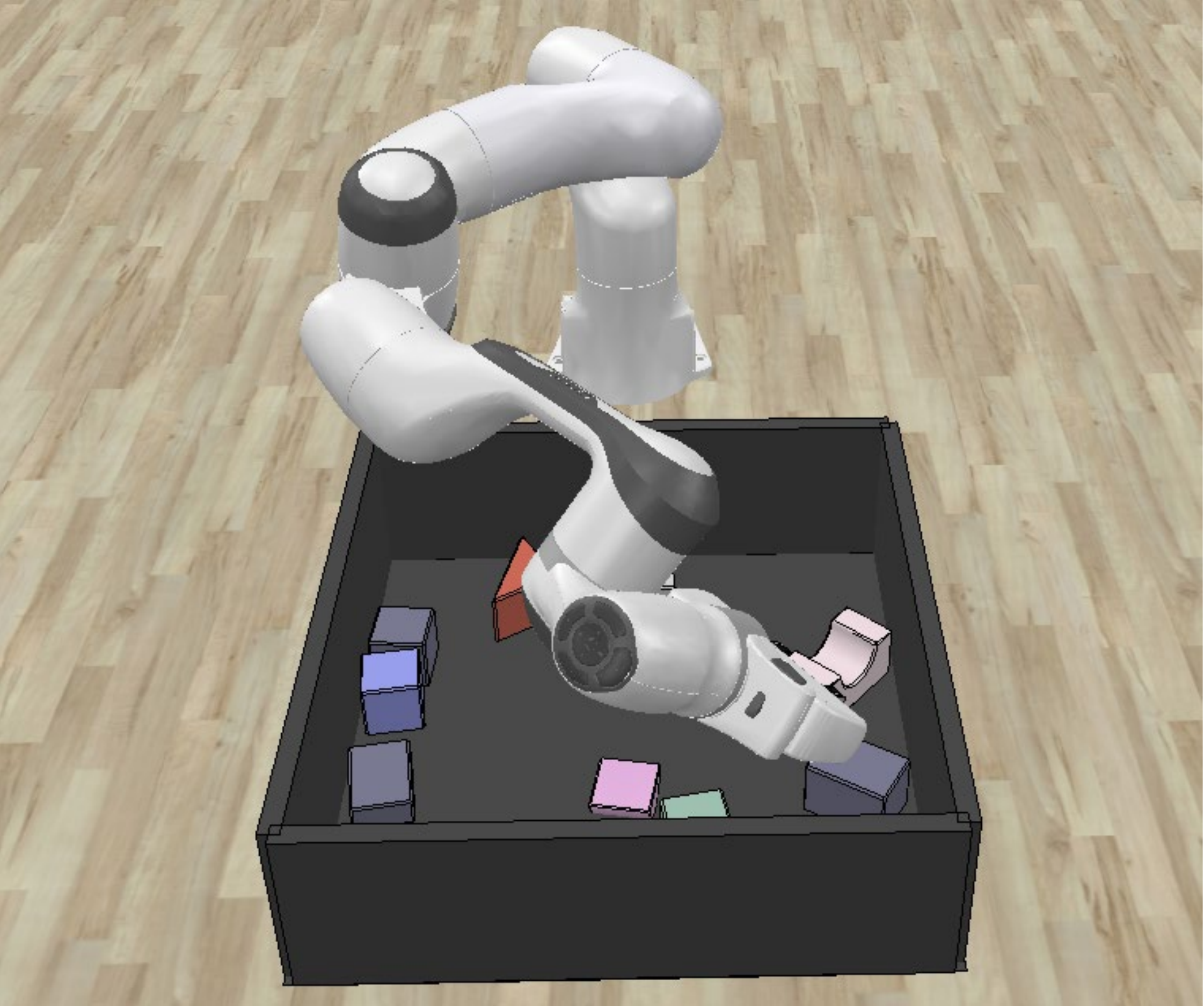}
    \caption{Challenging arrangement}
  \end{subfigure}
    \caption{\textbf{Different arrangements in simulation.} The standard arrangement in (a) test our approach in a common settings while the challenging arrangement in (b) reflects a manually designed scenario where collision-awareness is even more critical.}
  \label{fig:simulation}
  \vspace{-8pt}
\end{figure}

\textbf{Evaluation Metrics:} We define two metrics for both simulation and real-world evaluations, the planning rate and the grasping rate. The first metric is defined as the planning rate $ = \frac{\text{\# of successful plans}}{\text{\# of total trials}}$. A motion planning is considered successful only if the motion planning algorithm is able to find a valid trajectory for the robot arm without any collision. For grasping, the grasping rate is defined as $ \frac{\text{\# of successful grasps}}{\text{\# of proposed grasps}}$. A grasp is successful only if the robot gripper successfully reaches the object and lifts the object by 15 cm. Because the performance of each method varies depending on the object arrangement (i.e., an object is much more difficult to grasp if it is close to surrounding structures), we design a standard and a challenging arrangement to demonstrate the effectiveness of the CARP, shown in Fig.~\ref{fig:simulation}. If a target object is placed at an unreachable location (e.g., target objects fall at corners of the bin), we rearrange the object and continue the experiment.

\subsection{Perception Experiments}
The perception module is tested on known and novel objects that are randomly dropped into a bin. Note the novel objects include challenging ones that are similar in color but different in shape, as shown in Fig.~\ref{fig:novel}, which is non-trivial for RGB-based perception. We further test the robustness of our perception module against occlusion by hiding 25\% of the input images. Table I summarizes the performance of different perception modules using a small fine-tune dataset of 500 and a test dataset of 1000 image pairs. The results show the effectiveness of depth information. Our perception module outperforms the RGB only version by 35.07\% on novel objects and is robust to occlusion.

\begin{table}[t]
\caption{Target Recognition Accuracy}
\vspace{-6pt}
\label{tab:query}
\begin{center}
\begin{tabular}{|c||c||c||c||c|}
\hline
  & Known & Novel & Occl. & Novel-Occl.\\
\hline
RGB Siamese & 78.6 & 63.3 & 73.6 & 54.4\\
\hline
RGB-D Siamese (Ours) & \textbf{96.3} & \textbf{85.5} & \textbf{93.5} & \textbf{80.4}\\
\hline
\end{tabular}
\vspace{-8pt}
\end{center}
\end{table}

\subsection{Simulation Experiments}
Our simulation environment is built in CoppeliaSim \cite{rohmer2013v} with Bullet \cite{Coumans:2015:BPS:2776880.2792704} physics engine 2.83. The simulation setup uses a Panda robot arm, various workspace structures, and different test objects. We choose an eye-on-hand camera configuration to expand the field of view and minimize occlusion from surrounding structures.

\begin{figure}[t]
  \begin{subfigure}{0.154\textwidth}
    \includegraphics[width=\textwidth]{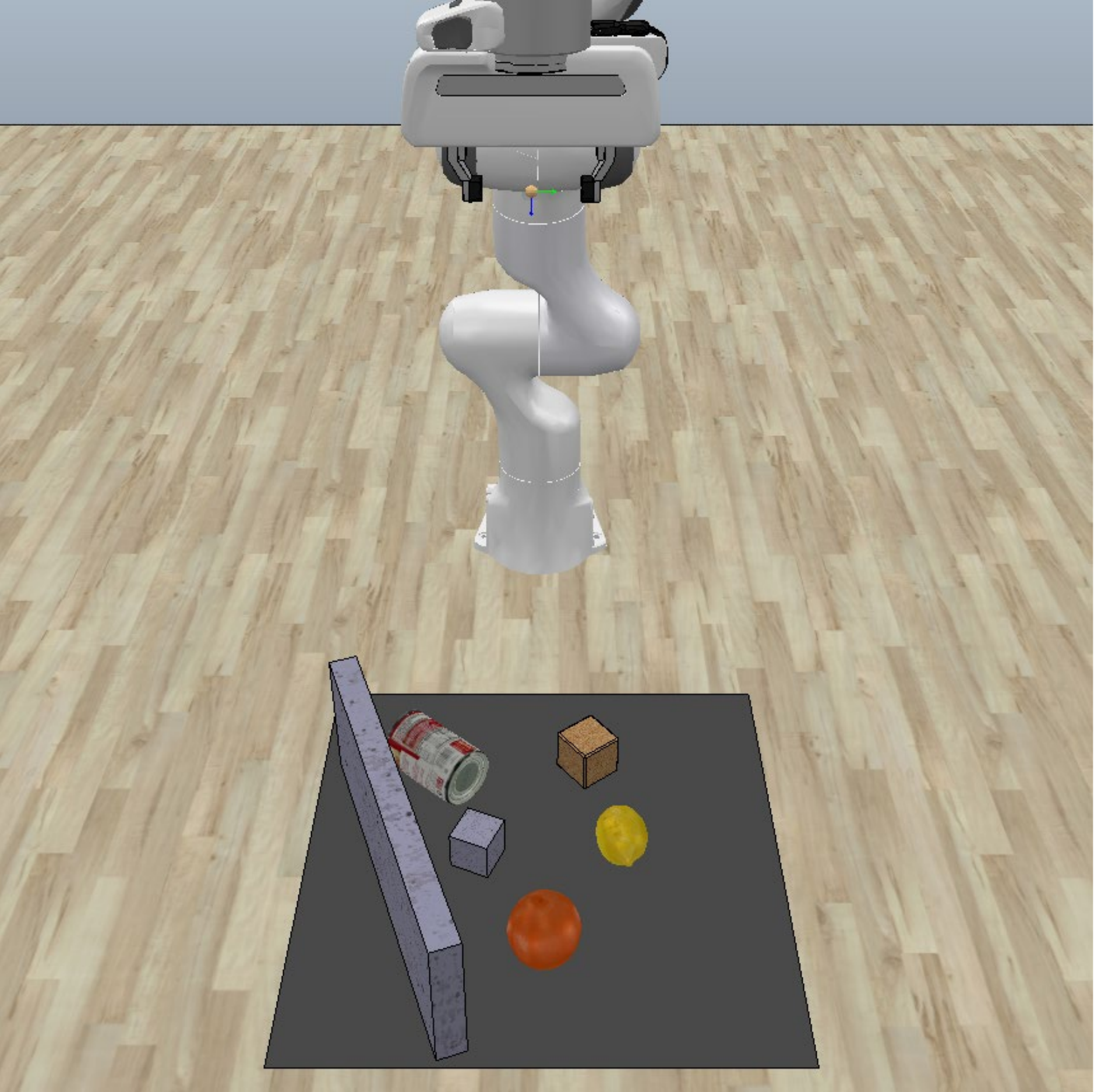}
    \caption{Wall}
  \end{subfigure}
  \begin{subfigure}{0.154\textwidth}
    \includegraphics[width=\textwidth]{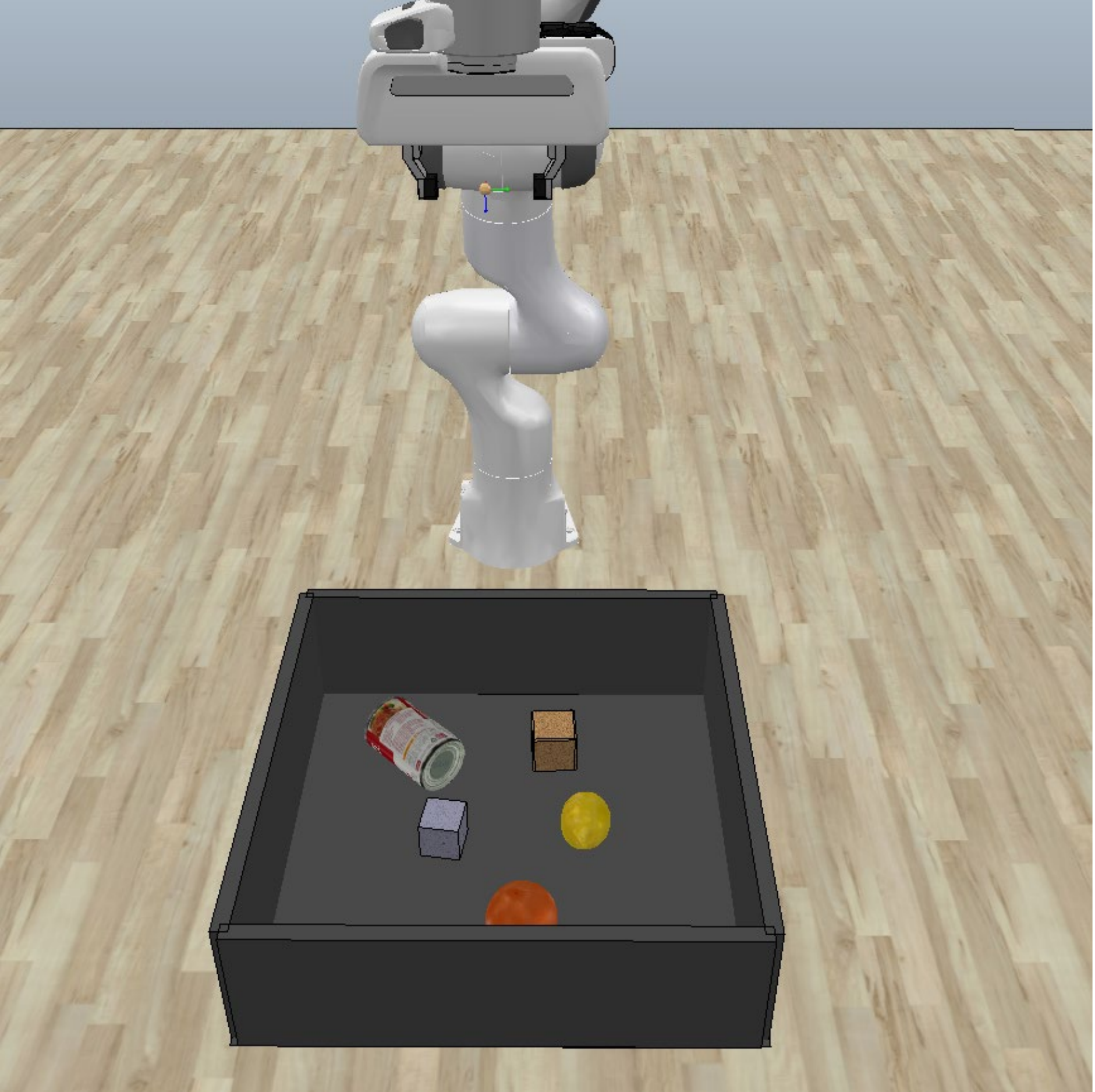}
    \caption{Large bin}
  \end{subfigure}
    \begin{subfigure}{0.154\textwidth}
    \includegraphics[width=\textwidth]{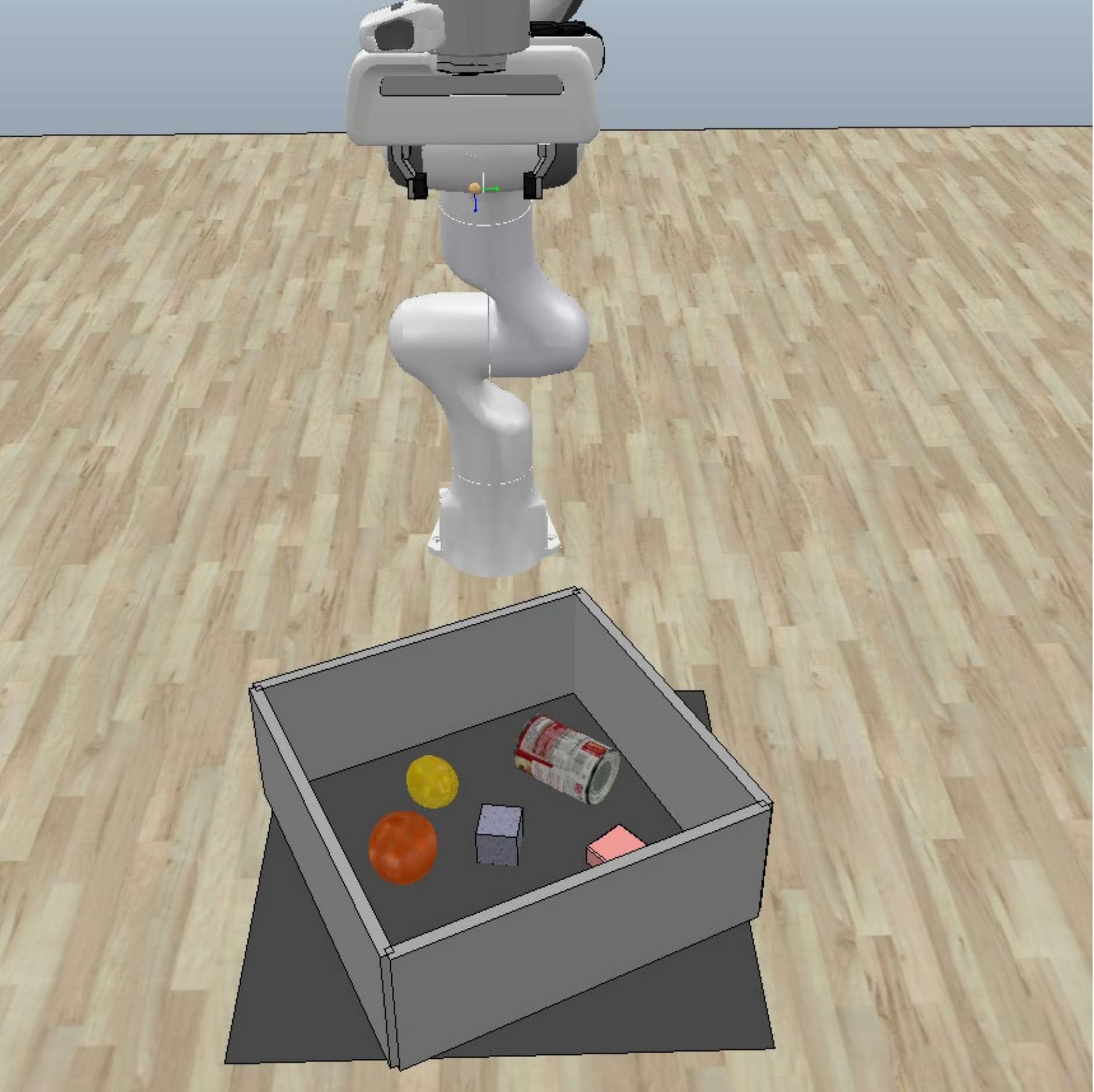}
    \caption{Small bin}
  \end{subfigure}
    \caption{\textbf{Different structures in simulation.} Our approach is able to grasp a target object in various surrounding structures with 6-DoF poses. As the testing environment gets increasingly challenging, the effectiveness of the CARP becomes more obvious.}
 \label{fig:sim_scenarios}
 \vspace{-8pt}
\end{figure}

\begin{figure}[t]
\centering
   \begin{subfigure}{0.49\textwidth}
       \centering
    \includegraphics[width=\textwidth]{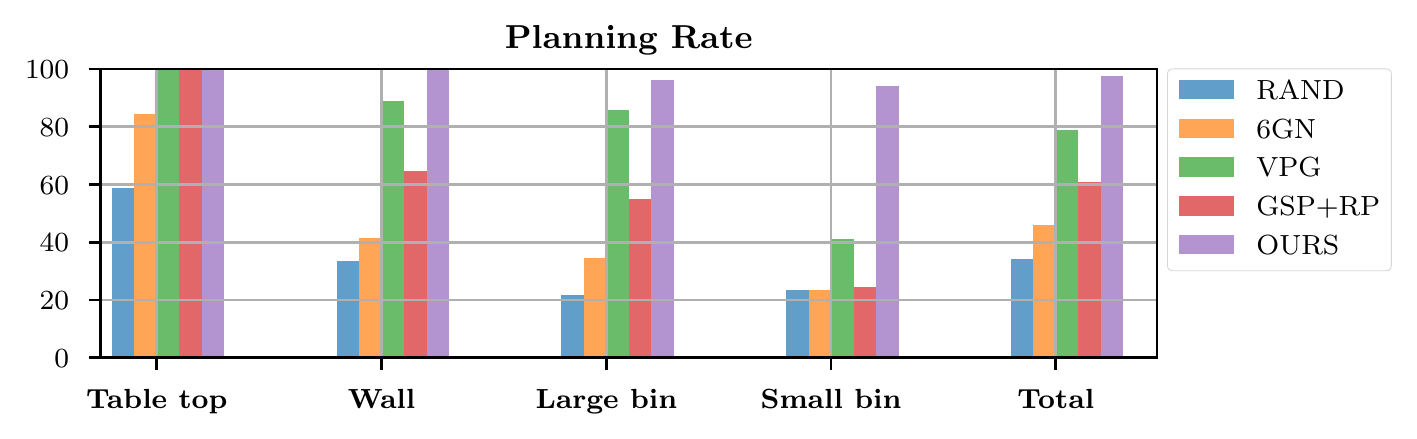}%
   \end{subfigure}
   \hspace{1em}
   \begin{subfigure}{0.49\textwidth}
       \centering
    \includegraphics[width=\textwidth]{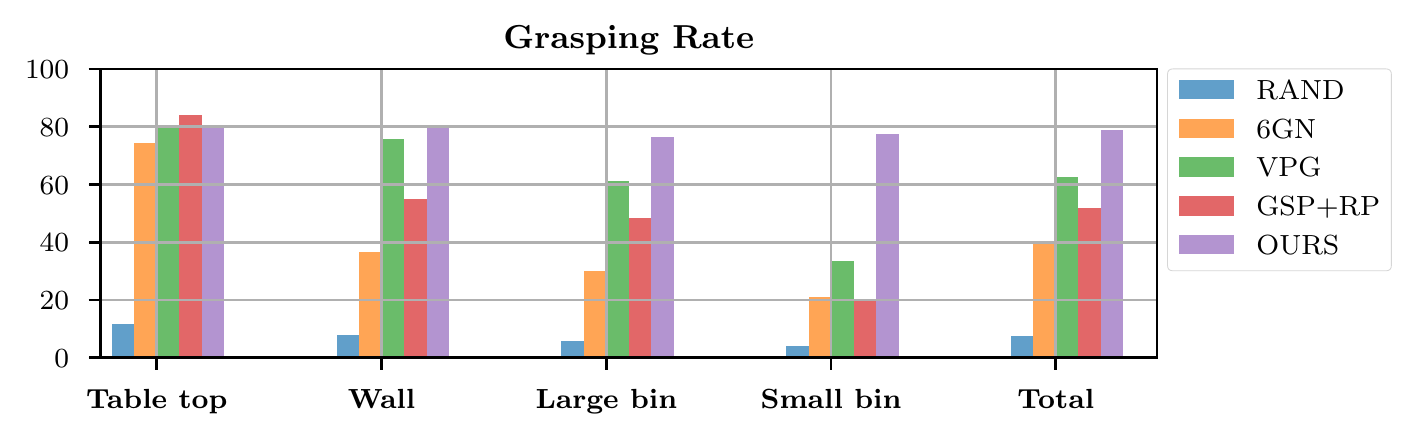}%
   \end{subfigure}
\caption{\textbf{Performance in simulation.} The planning rate (top) and grasping rate (bottom) of each approach in various surrounding structures. Our approach achieves 97.55\% planning rate and 78.78\% grasping rate on average, suggesting that it is the most effective.}
\label{fig:perf in sim}
\vspace{-8pt}
\end{figure}

In the standard arrangements, we use different workspace structures, such as walls, large boxes, and small boxes, and randomly orient them, as shown in Fig.~\ref{fig:sim_scenarios}. Ten objects are randomly dropped in the environment, and then we execute each approach 31 times. The target object is dropped into the workspace after a successful grasp to keep the test objects consistent. We report the results of each method in the standard arrangement in Fig.~\ref{fig:perf in sim}. We noticed that as the workspace's constraints become increasingly stringent, other approaches experience notable planning difficulties due to collision, while the planning rate of our approach degrades much less with the help of the CARP, hence increasing both the planning and grasping efficiencies. On average, our approach is able to achieve a 97.55\% planning rate and a 78.78\% grasping rate, which outperforms the baselines by large margins. The \textbf{\texttt{GSP+RP}} serves as an ablation baseline, introducing collision-awareness with the CARP increases both the planning and grasping rate of \textbf{\texttt{GSP+RP}} by more than 60\%. We notice that 6-DoF GraspNet suffers from infeasible grasp poses. Although \textbf{\texttt{VPG}} manages to avoid some of the planning challenges by using top-down grasping exclusively, it fails both planning and grasping tasks when the target object is close to the workspace structures (e.g., Small bin). 

In the challenging arrangement, the objects are manually dropped close to the peripheral of the bin. We evaluate the methods and summarize the results in Table~\ref{tab:challenging}. The evaluation of the challenging arrangement requires better reasoning about surrounding structures for grasping. Our approach outperforms the best-performing baseline by 127.31\% in grasping rate, showing that \textbf{\texttt{OURS}} is especially effective under this scenario.

\begin{table}[t]
\caption{Challenging Arrangement in Simulation}
\label{tab:challenging}
\begin{center}
\begin{tabular}{|c||c||c||c||c||c|}
\hline
  & RAND & VPG & 6GN & GSP+RP & OURS\\
\hline
Planning Rate & 12.90 & 41.94 & 21.57 & 26.67 & 93.55\\
\hline
Grasping Rate & 3.23 & 35.48 & 20.46 & 23.33 & 80.65\\
\hline
\end{tabular}
\end{center}
\end{table}

\subsection{Real-robot Experiments}
\begin{figure}[t]
  \begin{subfigure}{0.154\textwidth}
    \includegraphics[width=\textwidth]{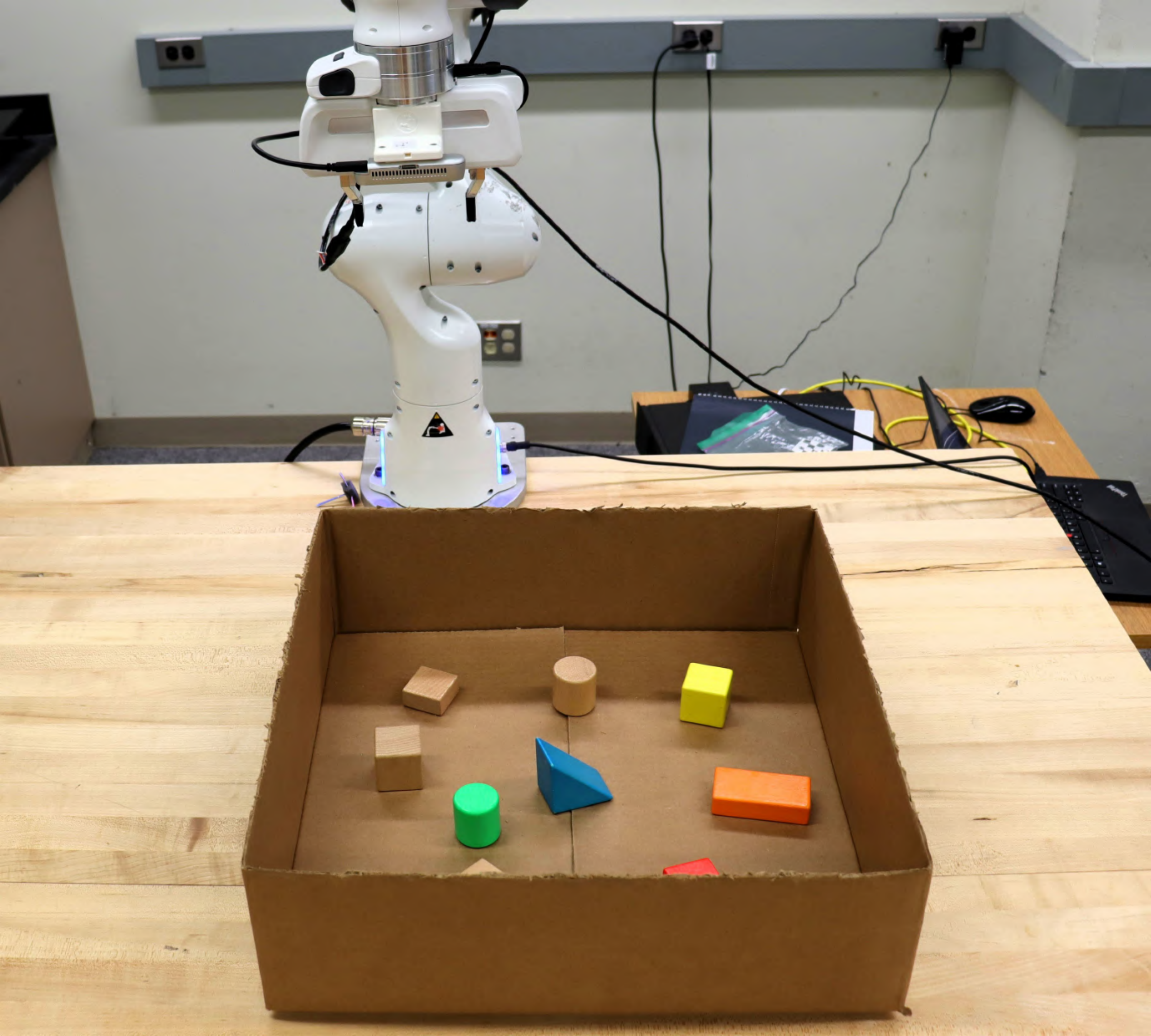}
    \par
    \vspace{0.1cm}
    \includegraphics[width=\textwidth]{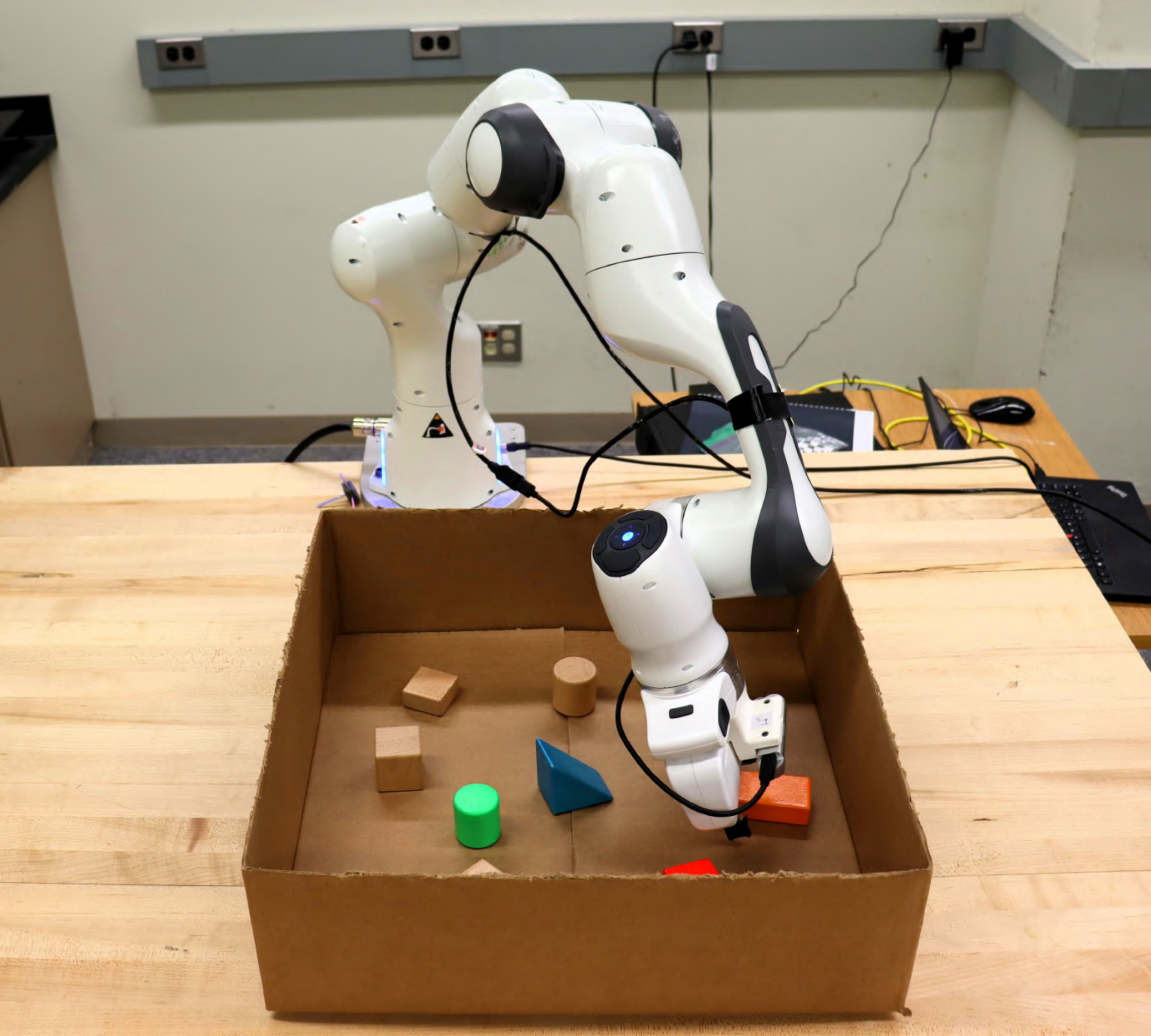}
    \caption{Standard}
  \end{subfigure}
  \begin{subfigure}{0.154\textwidth}
    \includegraphics[width=\textwidth]{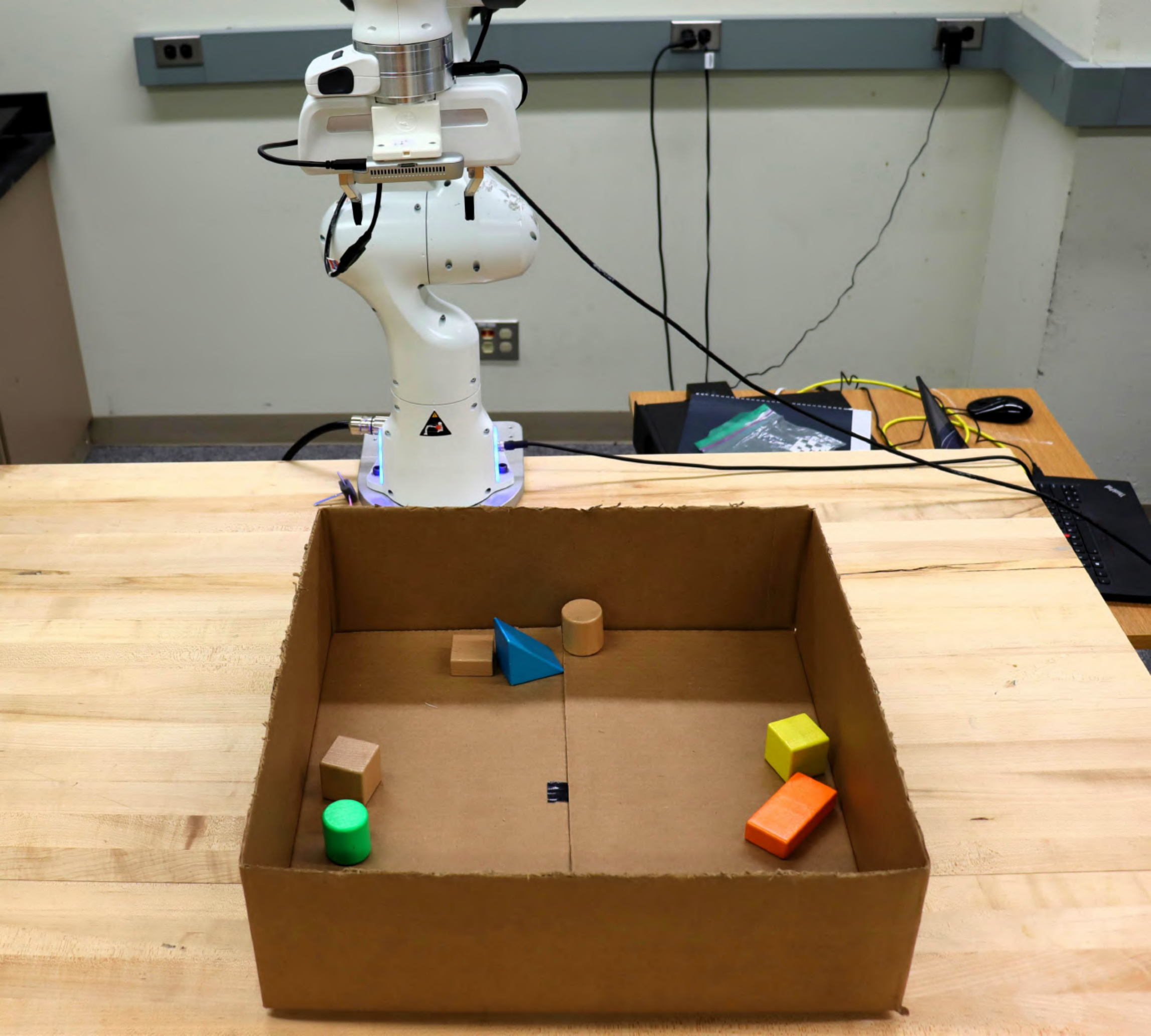}
    \par
    \vspace{0.1cm}
    \includegraphics[width=\textwidth]{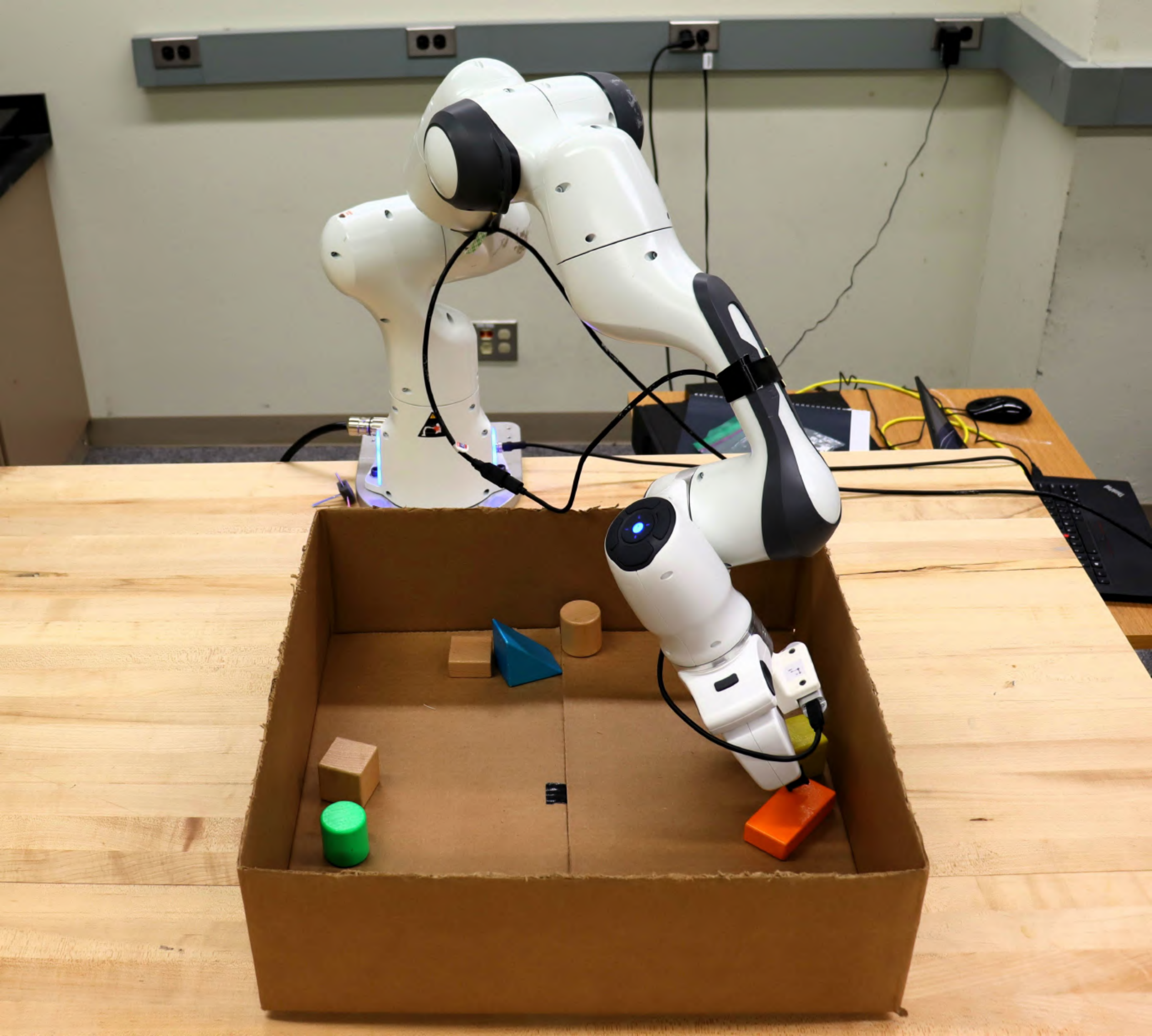}
    \caption{Challenging}
  \end{subfigure}
    \begin{subfigure}{0.154\textwidth}
    \includegraphics[width=\textwidth]{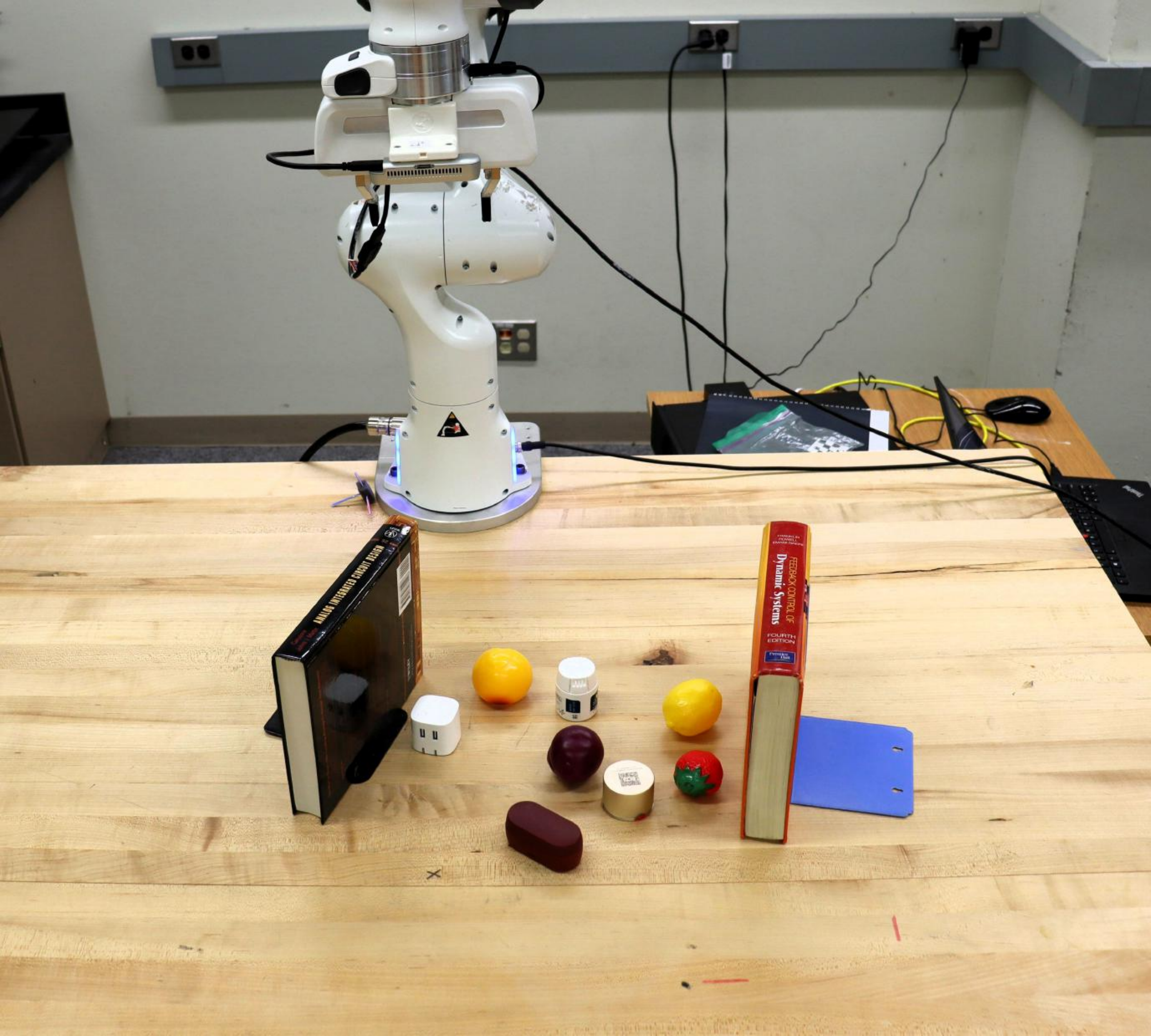}
    \par
    \vspace{0.1cm}
    \includegraphics[width=\textwidth]{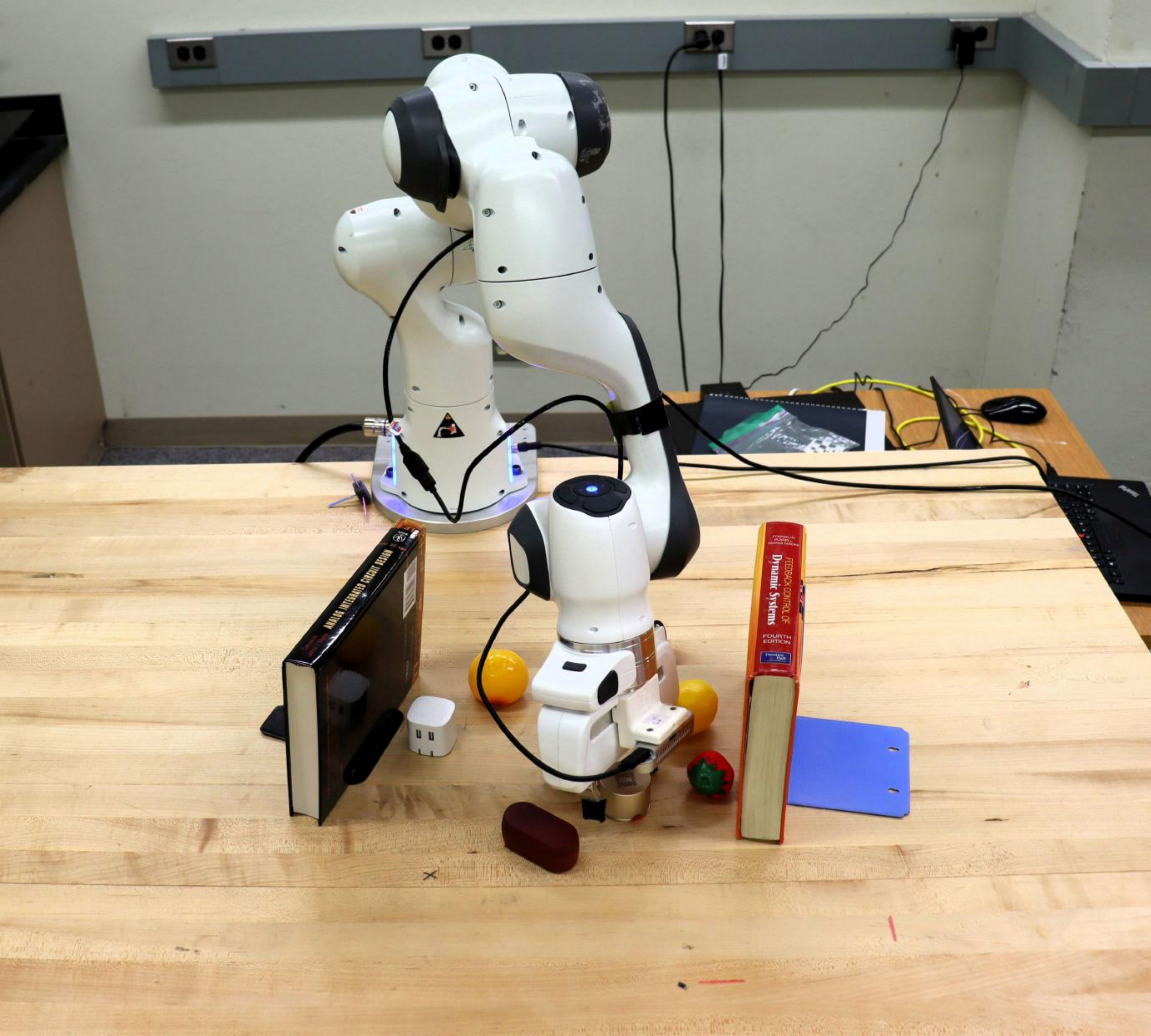}
    \caption{Novel}
  \end{subfigure}
    \caption{\textbf{Examples of real world experiments.} Our approach is able to grasp a target object that is close to the wall with 6-DoF grasp poses. This task is challenging as any small variation in grasp pose may lead to a collision with the surrounding structures.}
 \label{fig:real_scenarios}
\end{figure}

\begin{figure}[t]
\centering
   \begin{subfigure}{0.49\textwidth}
       \centering
    \includegraphics[width=\textwidth]{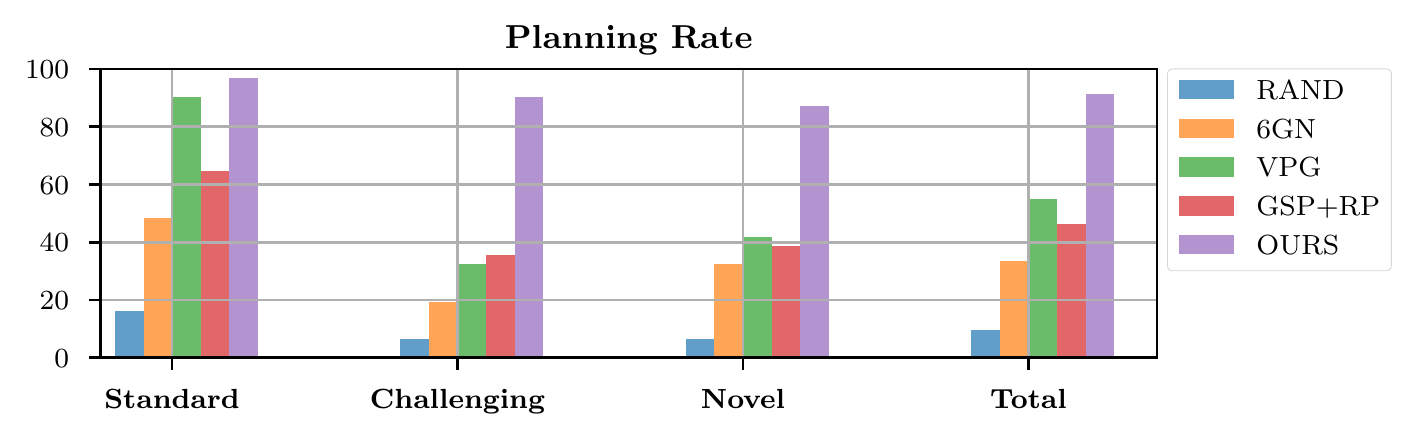}%
   \end{subfigure}
   \hspace{1em}
   \begin{subfigure}{0.49\textwidth}
       \centering
    \includegraphics[width=\textwidth]{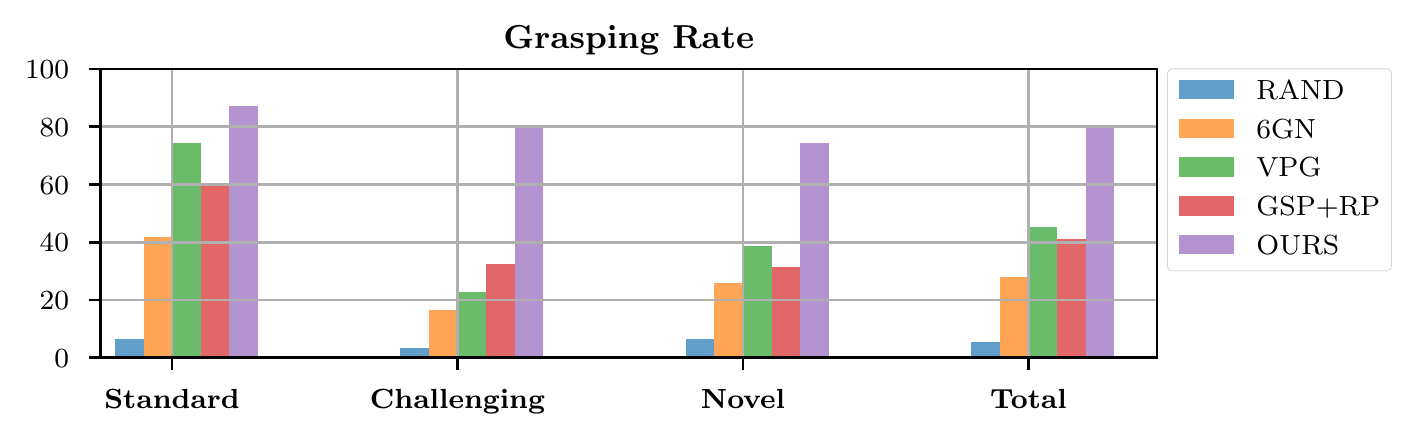}%
   \end{subfigure}
   \vspace{-8pt}
\caption{\textbf{Performance in the real world.} The planning rate (top) and grasping rate (bottom) of each approach in different arrangements. Our approach achieves 91.40\% planning rate and 80.65\% grasping rate on average, corroborating the simulation results.}
\label{fig:perf in real}
\end{figure}

We further evaluate our approach and the baselines on a Franka Emika Panda robot. A single-view RGB-D image is taken with an Intel Realsense D435 camera in eye-on-hand configuration. For a given pose, the robot follows the corresponding trajectory generated with MoveIt in open-loop. Fig.~\ref{fig:real_scenarios} shows our real-robot experiment settings, which include standard, challenging, and novel arrangements of objects in the real world. Following the same evaluation metrics, we compare our approach with \textbf{\texttt{RAND}}, \textbf{\texttt{6GN}}, \textbf{\texttt{VPG}},  and \textbf{\texttt{GSP+RP}}. Fig.~\ref{fig:perf in real} compiles the performance of each method in different scenarios for 31 runs.

Overall, our approach is able to perform consistently well in the real world and achieves the highest planning rate and grasping rate. Both \textbf{\texttt{GSP+RP}} and \textbf{\texttt{VPG}} are prone to predicting unreachable poses because they overlook the other environment information, which contributes to failed grasps. During testing, we realized the benefits as well as the limitations of our approach. The voxel grid of the CARP has a fixed resolution\footnote{One voxel of the CARP covers $2.5^3 \text{cm}^3$ of real robot workspace.}, and thus unable to capture smaller variations of either grasp poses or point cloud. In the very challenging cases where small variations can influence the collision-free probability (e.g., target object located side-by-side with the wall), the resolution limitation on the voxel degrades the performance.

\begin{figure}[t]
    \centering
    \includegraphics[width=\linewidth]{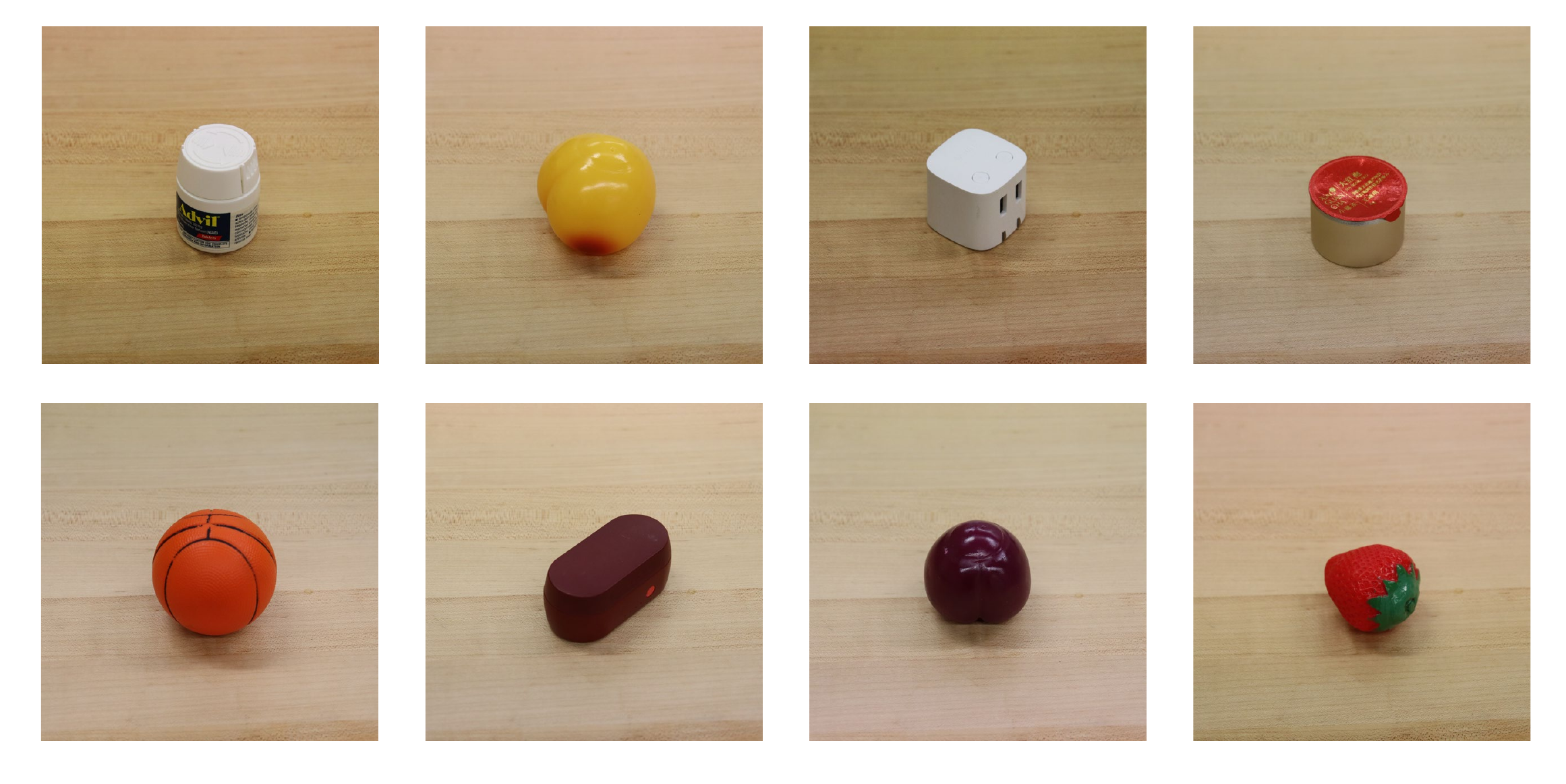}
    \vspace{-18pt}
    \caption{\textbf{Examples of novel objects.} We show that our work is able to generalize to objects that were not seen during training.}
    \label{fig:novel}
\end{figure}

Our approach generalizes well to novel environments. We test our system with a random collection of novel objects, as shown in Fig.~\ref{fig:novel}. Novel experiment results in Fig.~\ref{fig:perf in real} suggest our system is able to grasp novel objects as well. Note that there is no further training in both our perception and grasping algorithms as the perception pipeline generalizes to novel objects thanks to the Siamese network and the 3D CNN-based CARP and GSP, although additional fine-tuning would further improve the accuracy of our system. As the point clouds are transformed to gripper coordinates, changing the robot's gripper pose will subsequently modify its coordinate system and leave the performance of the system intact. The CARP is implicitly conditioned on robot hardware, therefore, it may require additional tuning in order to generalize to other robot manipulators. 

\section{CONCLUSION}
In this work, we presented the Collision-Aware Reachability Predictor (CARP), a learning-based approach that is able to accurately estimate collisions between the robot arm and surrounding structures using spatial information. Simulated and real experiments in various scenarios clearly showed the benefit of using the CARP in terms of planning rate and grasping rate. We further proposed a grasping pipeline that integrated our perception module and grasping module, and achieved, on average, a 91.40\% planning rate and a 80.65\% grasping rate in real-robot experiments. The proposed approach outperformed the other baseline methods by large margins. As future work, it would be interesting to include robot hand shapes in learning, so that our approach could better generalize to different robot hands.


\bibliographystyle{IEEEtran}
\bibliography{IEEEabrv,IEEEexample}
\end{document}